\documentclass[letterpaper]{article} 
\usepackage[submission]{aaai24}  
\usepackage{times}  
\usepackage{helvet}  
\usepackage{courier}  
\usepackage[hyphens]{url}  
\usepackage{graphicx} 
\usepackage{natbib}  
\usepackage{caption} 
\usepackage{algorithm}
\usepackage{algorithmic}
\usepackage{amsthm}

\usepackage{newfloat}
\usepackage{listings}
\DeclareCaptionStyle{ruled}{labelfont=normalfont,labelsep=colon,strut=off} 
\floatstyle{ruled}
\newfloat{listing}{tb}{lst}{}
\floatname{listing}{Listing}
\usepackage{amsmath}
\usepackage{amsfonts}
\title{Deep Evidential Learning for Bayesian Quantile Regression}

\author{
    Frederik Boe Hüttel\textsuperscript{\rm 1},
    Filipe Rodrigues\textsuperscript{\rm 1},
    Francisco C. Pereira\textsuperscript{\rm 1}.
}
\affiliations{
    \textsuperscript{\rm 1}Machine Learning for Smart Mobility, Transport Division, DTU Management \\
    Bygningstorvet 116B \\
    2800 Kgs. Lyngby, Denmark \\
    fbohy@dtu.dk

}

\begin{document}

\maketitle
\begin{abstract}
It is desirable to have accurate uncertainty estimation from a single deterministic forward-pass model, as traditional methods for uncertainty quantification are computationally expensive. 
However, this is difficult because single forward-pass models do not sample weights during inference and often make assumptions about the target distribution, such as assuming it is Gaussian.
This can be restrictive in regression tasks,
where the mean and standard deviation are inadequate to model the target distribution accurately.
This paper proposes a deep Bayesian quantile regression model that can estimate the quantiles of a continuous target distribution without the Gaussian assumption.
The proposed method is based on evidential learning, which allows the model to capture aleatoric and epistemic uncertainty with a single deterministic forward-pass model.
This makes the method efficient and scalable to large models and datasets. 
We demonstrate that the proposed method achieves calibrated uncertainties on non-Gaussian distributions, disentanglement of aleatoric and epistemic uncertainty, and robustness to out-of-distribution samples.


\end{abstract}

\section{Introduction}
\begin{figure}[t]
    \centering
    \includegraphics[width=0.9\columnwidth]{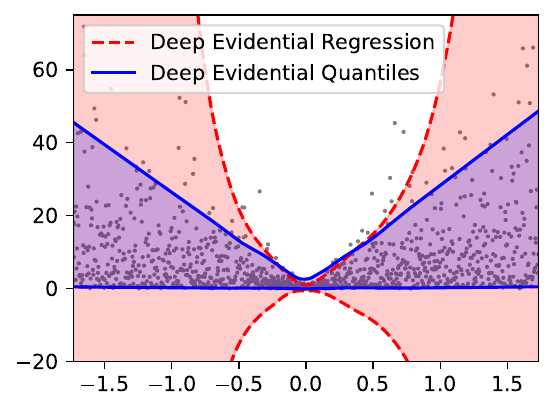}
    \caption{\textbf{Aleatoric Uncertainty} for our evidential Bayesian quantile regression trained on observations (black dots) with exponential noise.
    Deep Evidential regression models (red) predict a mean and variance, which do not fit asymmetrical noise distributions. 
    Our proposed approach (blue) does not have this limitation while benefiting from evidential models' epistemic uncertainty.
    Shaded areas correspond to a $90\%$ prediction interval.
    }
    \label{fig:figure_1}
\end{figure}
In machine learning, it is often essential to quantify the uncertainty of model predictions because it can provide valuable insights into the performance and limitations of the predictions.
Uncertainty can be broken down into two kinds:
\textit{aleatoric} and \textit{epistemic}.
Aleatoric results from uncertainty in the data, and epistemic is the inherent uncertainty of the model parameters \cite{kendall2017uncertainties}.

In regression tasks, aleatoric uncertainty can be modelled by assuming a parametric form of the target distribution and estimating the parameters of that distribution. For example, Gaussian aleatoric uncertainty can be estimated with a mean and variance, which can be learned with maximum likelihood \cite{nix1994estimating}. 
The mean of the distribution represents the predicted value of the target variable, while the variance represents the uncertainty about the target; together, these parameters are used to create prediction intervals for new test points.
In contrast, modelling epistemic uncertainty in modern neural networks (NNs) poses challenges.
The over-parameterised nature of modern NNs makes the functional form unsuitable for integration, rendering exact Bayesian inference intractable \cite{blundell2015weight}.
Various Bayesian NNs (BNNs) approaches have been proposed to address this issue. For example, using the reparametrization trick to sample weights from the weight posteriors \cite{blundell2015weight, kingma2015variational}, using dropout during testing to sample multiple weights \cite{gal2016dropout} or using Markov chain Monte Carlo methods to approximate posterior distributions over the weights given the data \cite{wilson2020bayesian}.
Deep ensembles are often considered the golden standard for modelling epistemic uncertainty, where multiple NNs are individually trained as the approximate distribution of the weights. 
The predictions are the mean of the ensemble, and the variance between them is the epistemic uncertainty \cite{lakshminarayanan2017simple, pearce2018high, ovadia2019can, pearce2020uncertainty}.
In addition to the limitation of using approximate inference to infer the posteriors, all these methods are limited by the computational cost associated with multiple forward passes, sampling and training of multiple large-scale neural networks.

Consequently, research has turned towards single-deterministic model approaches for the estimation of epistemic uncertainty \cite{van2020uncertainty, liu2020simple}. 
One approach is Deep Evidential Learning, which learns both aleatoric and epistemic uncertainty without sampling or training on out-of-distribution data~\cite{amini2020deep}.
Evidential deep learning places a prior over the parameters of the likelihood function instead of putting a prior over the weights.
This prior is a higher-order \emph{evidential} distribution, and samples from this distribution are instances of likelihood functions from the lower-order distribution.
Training a neural network to output the parameters of the evidential distributions leads to parametrised representations of epistemic and aleatoric uncertainty \cite{amini2020deep}.
This means that instead of learning the mean and variance of the target distribution, a NN is trained to learn the parameters of the evidential distribution \cite{ulmer2023prior}. 
These qualities make the model quite attractive from an application point of view, and much research has been done on the architecture and theoretical aspects of it \cite{meinert2021multivariate, Oh_Shin_2022, meinert2023unreasonable,yu2023evidentailneural}.

Despite the benefits of deep evidential regression, one of the main limitations is the reliance on Gaussian assumptions for the lower-order likelihood function, i.e., the aleatoric uncertainty.
This limits the modelling applications as the aleatoric uncertainty is constrained to mean and variance predictions, which might not describe the target distribution sufficiently. 
To address this limitation, we propose and derive an evidential Bayesian quantile regression model, where aleatoric uncertainty is not restricted to mean and variance but modelled with the quantiles of the target distribution instead. By adopting this approach, we retain the advantages of evidential learning in estimating epistemic uncertainty in a single forward pass while facilitating more flexible modelling of non-Gaussian aleatoric uncertainties - see Figure \ref{fig:figure_1} for an example.
Concretely, this paper makes the following contributions:
\begin{enumerate}
    \item Formulation and derivation of an evidential model and likelihood for quantile regression which can model non-Gaussian noise distributions.
    \item A scalable deep Bayesian quantile regression model without the need for sampling during inference.
    \item Evaluation of aleatoric and epistemic uncertainty on real-world datasets with comparisons with state-of-the-art BNN uncertainty estimation techniques.
\end{enumerate}

\section{Background}
\subsection{Quantile Regression}
We consider the following supervised regression problem, where given a dataset $\mathcal{D}$, with $N$ paired observations, $\mathcal{D}=\{\boldsymbol{x}_i,y_i\}_{i=1}^N$, we aim to learn a mapping $f$, parameterised by a set of weights $\boldsymbol{w}$, which approximately solves the following optimisation problem, 
\begin{equation}
    \min _{\boldsymbol{w}} J(\boldsymbol{w}) ; \quad J(\boldsymbol{w})=\frac{1}{N} \sum_{i=1}^N \mathcal{L}_i(\boldsymbol{w})\, ,
\end{equation} 
where $\mathcal{L}_i$ denotes the loss function, that describes how well $f(\boldsymbol{x}_i;\boldsymbol{w})$ fits the targets $y_i$. 
Unlike traditional regression problems, which minimise the mean squared error loss and thereby regress to the mean, we consider quantile regression, which instead learns a specified quantile $q$ of a target distribution \cite{koenerk1978regression}.
This is solved by using the tilted loss for a given quantile $q$,
\begin{equation}
\mathcal{L}_i(\boldsymbol{w})= \rho_q\left(\varepsilon_i\right) = \max(q \varepsilon_i, (q-1)\varepsilon_i)\, ,
\end{equation}
where $\varepsilon_i$ denotes the residual for observation $i$.
The entire target distribution (aleatoric uncertainty) can be modelled using multiple estimates of these quantiles.
However, the model does not explicitly model any uncertainties of the quantiles or the epistemic uncertainty of the model.

\subsection{Maximum Likelihood estimation}
A way to model the aleatoric uncertainty of the quantiles is with maximum likelihood estimation, where the objective is to learn the parameters of a distribution that maximise the likelihood of the training data. 
Traditional regression settings with Gaussian assumptions estimate a mean parameter ($\mu$) and variance parameter ($\sigma^2$) to parameterise the target distribution instead of a single-value prediction. 
For the case of quantile regression, the quantile follows an asymmetric Laplace distribution, with mean and variance parameter $\boldsymbol{\theta}_i=(\mu_i, \sigma_i)$, and an asymmetrical parameter equal to the quantile $q$ \cite{yu2005asym}.
We can infer the $\boldsymbol{\theta}_i$, which maximises the likelihood of the targets $y$, given $p(y_i|\boldsymbol{\theta}_i)$. This is done by minimising the negative log-likelihood function for the asymmetrical Laplace distribution \cite{Boukouvalas2012Gaussian}.  
\begin{equation}
\label{eq:laplace_log_like}
\begin{split}
\mathcal{L}_i(\boldsymbol{w}) &= -\log p\left(y_i \mid \boldsymbol{\theta}_i, q \right) \\
&=- \log \left(\frac{q(1-q)}{\sigma_i} \exp \left(-\rho_q \left(\frac{y_i-\mu_i}{\sigma_i}\right)\right) \right)
\end{split}
\end{equation}
By predicting $\boldsymbol{\theta}_i$, $\mu_i$ becomes the prediction of the quantile, and $\sigma_i$ denotes the aleatoric uncertainty of the quantile. 

\subsection{Bayesian Quantile regression}
It is often infeasible and intractable to perform Bayesian inference directly on Equation~\ref{eq:laplace_log_like} to get epistemic uncertainties \cite{tsionas2003bayesian, dai2023high}.
Alternatively, the likelihood function in Equation \ref{eq:laplace_log_like} can be reformulated as a scalar mixture of Gaussians \cite{kotz2001laplace, kozumi2011gibbs} such that,
\begin{equation}
\label{eq:scalar_mixture_model}
    p(y_i \mid \boldsymbol{\theta}_i, z_i, q) =  \mathcal{N}\left(\mu_i+\tau z_i, \omega \sigma_i z_i\right) ,
\end{equation}
where $\tau = \frac{1-2q}{q(1-q)}$ and $\omega=\frac{2}{q(1-q)}$ are quantile-specific constants and $z_i \sim \operatorname{Exp}\left(\frac{1}{\sigma_i}\right)$.

This formulation of the quantile regression problem has two advantages: \textbf{1)} it allows for Bayesian inference of the quantiles, such as variational inference \cite{Abeywardana_Ramos_2015}, or Gibbs sampling \cite{kozumi2011gibbs}, and \textbf{2)} even though the distribution of $y_i$ is Gaussian, $\mu_i$ can still model quantiles of non-Gaussian distributions. 
We leverage these two advantages to extend the deep evidential regression models to Bayesian quantile regression, such that evidential models can model non-Gaussian distributions.

\section{Evidential Uncertainty for Quantile Regression}
We consider a hierarchical Bayesian structure of the regression problem.
We assume that the observed targets, $y_i$, come from a Gaussian distribution parameterised as Equation~\ref{eq:scalar_mixture_model}, but \textit{$\mu_i$ and $\sigma_i$ are unknown}.
The idea is that for a given quantile $q$, the mean of $y_i$ follows the corresponding quantile value, $\mu_i$, offset by a quantile specific constant, $\tau$, scaled by the aleatoric uncertainty of the quantile ($z_i \sim \operatorname{Exp}\left(\frac{1}{\sigma_i}\right)$).
Since the values of $\mu_i$ and $\sigma_i$ are unknown, we seek to estimate the distribution of these parameters.
We place a Gaussian prior on the unknown $\mu_i$ and an Inverse-Gamma prior on the unknown $\sigma_i$~\cite{Abeywardana_Ramos_2015, chu2023bayesian},
\begin{equation}
    \begin{gathered}
y_i \sim \mathcal{N}\left(\mu_i+\tau z_i, \omega \sigma_i z_i\right) \\
\mu_i \sim \mathcal{N}\left(\gamma_i, \sigma_i \nu_i^{-1}\right) \quad \sigma_i \sim \Gamma^{-1}(\alpha_i, \beta_i) .
\end{gathered}
\end{equation}
where $\Gamma^{-1}(\cdot)$ is the inverse gamma distribution and $\gamma_i \in \mathcal{R}$, $\nu_i > 0$, $\alpha_i > 1$, $\beta_i > 0$ and $z_i \sim \operatorname{Exp}\left(\frac{1}{\beta_i/\left(\alpha_i-1\right)}\right)$.
We denote the set $\boldsymbol{m}_i = (\gamma_i, \nu_i, \alpha_i, \beta_i)$.
Together the distributions of $\mu_i$ and $\sigma_i$ form the Normal-Inverse-Gamma (NIG) evidential prior \cite{bishop2006pattern, amini2020deep},
\begin{equation}
\begin{aligned}
p(\mu_i, \sigma_i \mid \boldsymbol{m}_i)  &=p(\mu_i\mid \boldsymbol{m}_i) p\left(\sigma_i\mid \boldsymbol{m}_i\right) \\
&=\mathcal{N}\left(\gamma_i, \sigma_i \nu_i^{-1}\right) \Gamma^{-1}(\alpha_i, \beta_i) \\
 =\frac{\beta_i^{\alpha_i} \sqrt{\nu_i}}{\Gamma(\alpha_i) \sqrt{2 \pi \sigma_i}} & \left(\frac{1}{\sigma_i}\right)^{\alpha_i+1} \exp \left\{-\frac{2 \beta_i+\nu_i(\gamma_i-\mu_i)^2}{2 \sigma}\right\}.
\end{aligned}
\end{equation}
The objective is to use a single deterministic NN to infer the parameters of this evidential distribution, namely the set $\boldsymbol{m}_i$.

\subsection{The Evidential Distribution}
Inference of the parameters $\boldsymbol{m}_i$ can be accomplished by maximizing the probability $p(y_i \mid \boldsymbol{m}_i)$.
This is done by marginalising over all the likelihood parameters $\mu_i$ and $\sigma_i$\footnote{We omit the dependency of $q$ for readability.},
\begin{equation}
\small
        p\left(y_i \mid \boldsymbol{m}_i\right)= \int_{\sigma_i=0}^{\infty} \int_{\mu_i=-\infty}^{\infty} p\left(y_i \mid \mu_i, \sigma_i \right) p\left(\mu_i, \sigma_i \mid \boldsymbol{m}_i\right) \mathrm{d} \mu_i \mathrm{d} \sigma_i.
\end{equation}


By placing the NIG prior on the likelihood parameters, there exists an analytical solution to this integral that produces a Student-t predictive distribution given by,
\begin{equation}
    p(y_i|\boldsymbol{m}_i) = \text{St}(y_i;\gamma_i+ \tau z_i, \frac{2\beta_i(1+\omega\nu_i z_i)}{\nu_i \alpha_i}, 2\alpha_i) \, .
\end{equation}
where $\text{St}(y; \mu_\text{st}, \sigma_\text{st}, v_\text{st})$ is the Student t-distribution evaluated at location $\mu_\text{st}$ and scale $\sigma_\text{st}$, with degrees of freedom $v_\text{st}$.

The mean of the t-distribution is equal to the mean of the scalar-mixture representation of the quantile regression problem (Equation \ref{eq:scalar_mixture_model}), while the scale is a combination of the quantile specific constant $\omega$ and the parameters of $\boldsymbol{m}_i$.
The complete derivation is provided in Appendix \ref{sec:deriviation}.

\subsection{Learning the evidential quantile distribution}
To learn the parameters $\boldsymbol{m}_i$ of the evidential distribution, we maximise the likelihood of the student t-distribution \cite{amini2020deep}.
The model is trained to minimise the negative log-likelihood, 
\begin{equation}
\begin{split}
    \mathcal{L}_i^{\mathrm{NLL}}&(\boldsymbol{w})=\frac{1}{2} \log \left(\frac{\pi}{\nu_i}\right)-\alpha_i \log (\Omega)+\left(\alpha_i+\frac{1}{2}\right) \\ 
     & \log \left(\left(y_i- \gamma_i+\tau z_i\right)^2      \nu_i+\Omega\right)+\log \left(\frac{\Gamma(\alpha_i)}{\Gamma\left(\alpha_i+\frac{1}{2}\right)}\right),
    \end{split}
\end{equation}
where $\Omega = 4\beta_i(1+\tau z_i \nu_i)$ and $z$ is the mean of the Exponential distribution of $z_i = \frac{\beta_i}{\alpha_i-1}$.

\subsection{Aleatoric and Epistemic Uncertainties}
As a result, the prediction of a quantile from the t-distribution is,
\begin{equation}
\hat{\mu}_i=\mathbb{E}_{p\left(\mu_i \mid \boldsymbol{m}\right)}\left[\mu_i\right]=\int \mu_i p\left(y_i \mid \boldsymbol{m}_i\right) \mathrm{d} \mu_i=\gamma_i
\end{equation}
Based on the evidential parameters $\boldsymbol{m_i}$, the NIG distribution is fully characterised. It allows us to parametrise the epistemic ($\operatorname{Var}[\mu_i]$), and the aleatoric uncertainty ($\mathbb{E}[\sigma_i]$), 
\begin{equation}
\underbrace{\mathbb{E}\left[\sigma_i\right]=\frac{\beta_i}{\alpha_i-1}}_{\text {aleatoric }}, \quad \underbrace{\operatorname{Var}[\mu_i]=\frac{\beta_i}{\nu_i(\alpha_i-1)}}_{\text {epistemic }} 
\end{equation}

\subsection{Minimising the evidence on errors}
We need to regularise training by applying an evidence penalty to minimise evidence of incorrect predictions \cite{amini2020deep}.
The parameters $\nu$ and $\alpha$ of the NIG prior can be interpreted as evidence to quantify the confidence on the prior of the mean.
However, $\beta$ denotes the variance of the model, and large values of $\beta$ lead to low confidence in the model's predictions, which implies a lack of evidence \cite{yu2023evidentailneural}. 
By using all the evidence parameters, we quantify the model confidence as,
\begin{equation}
    \Phi_i = 2\nu_i +\alpha_i + \frac{1}{\beta_i}.
\end{equation}

\begin{figure*}[t]
\centering
\includegraphics[width=0.9\columnwidth]{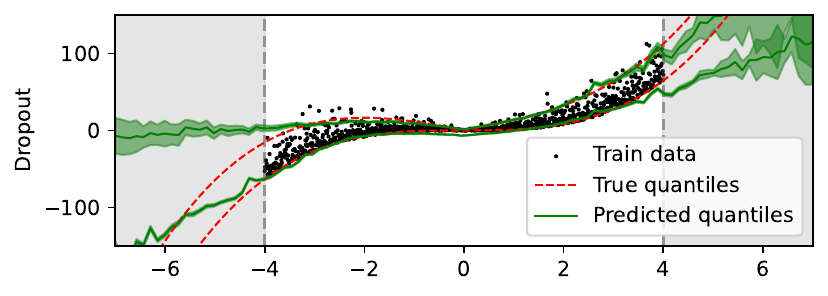}
\includegraphics[width=0.9\columnwidth]{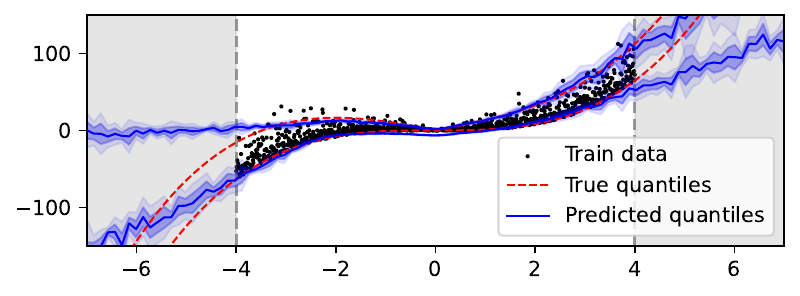}
\includegraphics[width=0.9\columnwidth]{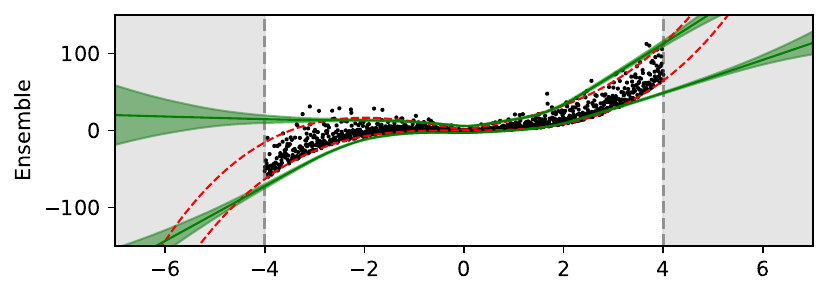}
\includegraphics[width=0.9\columnwidth]{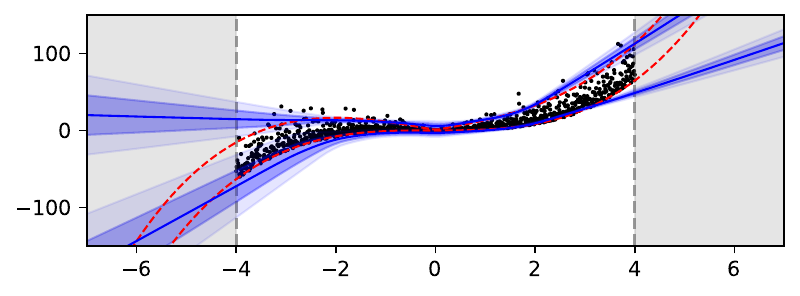}
\includegraphics[width=0.9\columnwidth]{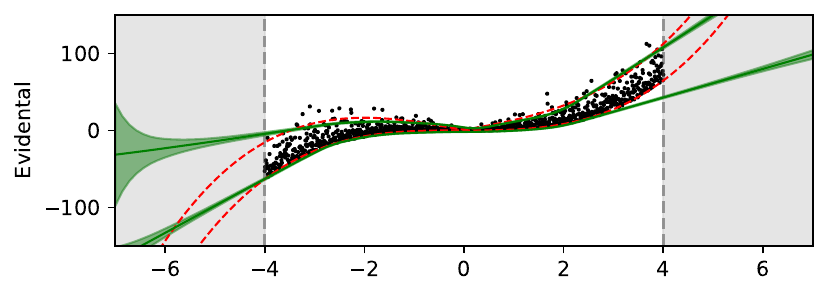}
\includegraphics[width=0.9\columnwidth]{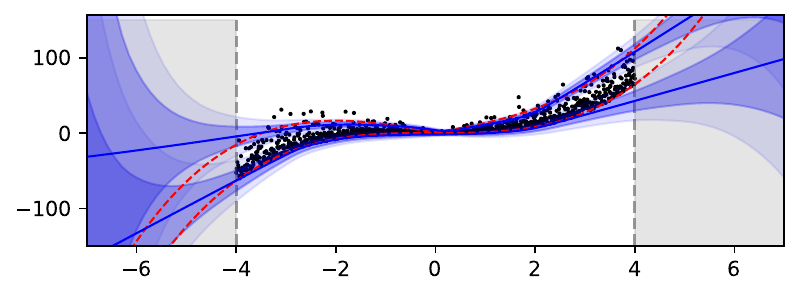}
\caption{\textbf{Uncertainty Estimation.} Aleatoric (left) and Epistemic (right) uncertainty estimation on the dataset $y=x^3+\varepsilon, \, \varepsilon \sim \operatorname{Exp}\left(\frac{1}{4|x|+0.2}\right)$, where $x\in [-4; 4]$, for the Dropout sampling, ensemble models and deep evidential quantiles.
All methods accurately model the quantiles within the training distribution, and epistemic uncertainty grows on OOD data.
}
\label{fig:quali_eli}
\end{figure*}
This encourages the model to output low confidence when the predictions are incorrect. 
Traditionally, this evidence has been scaled with the absolute error \cite{amini2020deep, yu2023evidentailneural}, however, since we consider quantile regression, we propose to scale the model evidence with the tilted loss instead,
\begin{equation}
    \mathcal{L}_i^{\mathrm{R}}(\boldsymbol{w})=\rho_q(y_i-\mathbb{E}\left[\mu_i\right]) \cdot \Phi_i=\rho_q(y_i-\gamma_i) \cdot \Phi_i\, .
\end{equation}
The tilted loss is a generalised version of the absolute error with a tilting parameter $q$. 
When the $q=0.5$, the titled loss equals the absolute error function \cite{rodrigues2020beyond}.
For high quantiles ($q>0.5$), an underestimation is penalized less than an overestimation and vice versa for the low quantiles ($q<0.5$).
This means the model is trained to output low values of $\nu$ and $\alpha$ and high values of $\beta$ (i.e. more confident predictions) when the high quantiles are overestimated and less when underestimated.


\paragraph{Summary and Implementation}
To summarise, the total loss of the model $\mathcal{L}_i(\boldsymbol{w})$ consists of the loss from the two terms, 1) for maximising the model fit and 2) for regularising the evidence.
The regulariser is scaled by a regularisation coefficient $\lambda$, 
\begin{equation}
\label{eq:total_loss}
    \mathcal{L}_i(\boldsymbol{w})=\mathcal{L}_i^{\mathrm{NLL}}(\boldsymbol{w})+\lambda \mathcal{L}_i^{\mathrm{R}}(\boldsymbol{w}) .
\end{equation}
The coefficient $\lambda$ trades off uncertainty inflation with the model fit \cite{amini2020deep}.
In practice, we train a NN to output the parameters of the evidential distribution $\boldsymbol{m}_i=f(\boldsymbol{x}_i;\boldsymbol{w})$. $\boldsymbol{m}_i$ is composed of 4 parameters, so $f$ has an output neuron for each parameter. The constraints on the parameters are enforced with the softplus activation function on the parameters, $\nu_i$, $\beta_i$ and $\alpha_i$.
In addition, we add 1 to $\alpha_i$ to satisfy the constraint that $\alpha_i > 1$.

Our approach presents a distinct advantage compared to previous work by learning the quantiles of the target distribution instead of a mean and standard deviation.
Therefore, we can extend the aleatoric uncertainty estimation to non-Gaussian distributions while obtaining parameterised epistemic uncertainty. 
In addition, it is common in quantile regression to estimate multiple quantiles at once, which our approach can easily achieve by increasing the number of evidential outputs for a model.

\section{Experiments}
\paragraph{Qualitative evaluation}
We begin our empirical analysis by comparing the qualitative performance of our proposed method.
We compare the uncertainty estimation visually with the following baselines,  Dropout sampling \cite{gal2016dropout} and Deep Ensembles \cite{lakshminarayanan2017simple} trained on the loss function for the asymmetrical Laplace distribution (Equation \ref{eq:laplace_log_like}). 
We use $n_\text{samples}=5$ samples of weights during inference for the baseline models.
We train the models on the dataset $y=x^3+\varepsilon, \, \varepsilon \sim \operatorname{Exp}\left(\frac{1}{4|x|+0.2}\right)$, where $x\in [-4; 4]$.
As is common in quantile regression, we estimate multiple quantiles simultaneously, namely the 5th and 95th quantiles. 
We compare the aleatoric and epistemic uncertainty estimation for the different models in Figure \ref{fig:quali_eli}. 
All methods capture the aleatoric uncertainty within the training distribution. As we test the models on out-of-distribution data, our method estimates the uncertainty accurately, with increasing epistemic uncertainty, as we move further away from the training distribution without the need for multiple forward passes during inference.
We find that a regularisation strength of $\lambda=0.5$ works well across the different model architectures and experiments. 
The complete training details for \emph{all} experiments are in Appendix \ref{app:experiments}.

\paragraph{Beyond the Gaussian distribution}
We test the proposed approach to model quantiles of non-Gaussian distributions. 
The baselines remain the same. However, we also include previous work on evidential regression that assumes that the target follows a Gaussian distribution \cite{amini2020deep}. Using the Gaussian output, we can compute the quantiles of the target distribution using the estimated mean and variance.
We generate synthetic data following $y=x^3 + \varepsilon$, where the noise distribution of $\varepsilon$ is non-Gaussian. 
Since we know the noise distribution, we can analytically compute their quantiles and compare them with the model's predictions. We compare the methods on the mean absolute error (MAE) to these quantiles.

Table \ref{tab:non_gaussian_results} shows the MAE between the estimated and theoretical quantiles. It shows that when the Gaussian assumption holds, all the models approximate the quantiles equally well, however as we deviate from the Gaussian assumption, we see that our proposed method is similarly good at modelling the quantiles compared to the dropout and ensembles as the MAE is comparable across the models. 
The performance of the evidential Gaussian model decreases as the Gaussian assumption is violated.

\begin{table}
    \centering
\resizebox{\columnwidth}{!}{
    \begin{tabular}{l|rrrr} 
 & \multicolumn{1}{c}{5th quantile} &  \multicolumn{1}{c}{25th quantile} &\multicolumn{1}{c}{75th quantile} &\multicolumn{1}{c}{95th quantile} \\ 
\hline & \multicolumn{4}{c}{Gaussian $\sim \mathcal{N}\left(0, 3|x|+0.2\right)$}  \\ \hline
Evidential Gaussian & $\mathbf{3.93 \pm 0.39}$& $\mathbf{2.73 \pm 0.03}$& $\mathbf{2.44 \pm 0.39}$& $\mathbf{4.08 \pm 0.15}$ \\
Dropout & $4.55 \pm 1.20$& $3.26 \pm 0.65$& $3.09 \pm 0.69$& $\mathbf{3.78 \pm 0.82}$ \\
Ensemble & $\mathbf{3.62 \pm 0.71}$& $\underline{\mathbf{2.44 \pm 0.50}}$& $\underline{\mathbf{2.28 \pm 0.33}}$& $\mathbf{3.06 \pm 0.64}$ \\
Evidential Quantile & $\underline{\mathbf{3.39 \pm 0.98}}$& $\mathbf{2.92 \pm 0.73}$& $\mathbf{2.51 \pm 0.78}$& $\underline{\mathbf{3.04 \pm 1.04}}$ \\
\hline & \multicolumn{4}{c}{Exponential $\sim \operatorname{Exp}\left(\frac{1}{4|x|+0.2} \right)$}  \\ \hline
Evidential Gaussian & $25.17 \pm 9.72$& $9.90 \pm 4.02$& $5.59 \pm 3.84$& $14.02 \pm 9.10$ \\
Dropout & $\mathbf{3.90 \pm 1.52}$& $\mathbf{2.82 \pm 0.70}$& $\mathbf{3.18 \pm 0.88}$& $4.83 \pm 1.21$ \\
Ensemble & $ \underline{\mathbf{3.11 \pm 0.67}}$& $\underline{\mathbf{2.00 \pm 0.15}}$& $\mathbf{2.55 \pm 0.40}$& $\mathbf{4.07 \pm 0.63}$ \\
Evidential Quantile & $\mathbf{3.71 \pm 1.80}$& $\mathbf{2.72 \pm 0.77}$& $\underline{\mathbf{2.52 \pm 0.44}}$& $\underline{\mathbf{3.34 \pm 0.64}}$ \\
\hline & \multicolumn{4}{c}{Gamma $ \sim \operatorname{Gamma}\left(3|x|, \frac{1}{2|x|+0.2}\right)$}  \\ \hline
Evidential Gaussian & $4.28 \pm 0.71$& $\mathbf{3.54 \pm 0.55}$& $5.43 \pm 0.65$& $10.34 \pm 0.69$ \\
Dropout & $3.75 \pm 0.61$& $\mathbf{3.48 \pm 0.54}$& $\mathbf{4.84 \pm 1.15}$& $7.76 \pm 2.00$ \\
Ensemble & $\mathbf{3.05 \pm 0.39}$& $\underline{\mathbf{2.64 \pm 0.19}}$& $\mathbf{3.42 \pm 0.76}$& $\mathbf{5.16 \pm 1.35}$ \\
Evidential Quantile & $\underline{\mathbf{2.89 \pm 0.50}}$& $\mathbf{3.24 \pm 0.99}$& $\underline{\mathbf{3.34 \pm 0.89}}$& $\underline{\mathbf{4.64 \pm 1.19}}$ \\
\hline & \multicolumn{4}{c}{Laplace $\sim \operatorname{Laplace}\left(0, 5|x|+0.2\right)$}  \\ \hline
Evidential Gaussian & $9.08 \pm 0.28$& $5.90 \pm 0.82$& $4.41 \pm 1.56$& $6.81 \pm 0.41$ \\
Dropout & $6.32 \pm 1.92$& $4.44 \pm 1.37$ &  $\mathbf{3.72 \pm 1.11}$& $5.28 \pm 1.84$ \\
Ensemble & $\underline{\mathbf{4.54 \pm 0.65}}$& $\underline{\mathbf{3.20 \pm 0.65}}$& $\underline{\mathbf{3.28 \pm 0.48}}$& $\mathbf{4.21 \pm 0.81}$ \\
Evidential Quantile & $\mathbf{4.47 \pm 0.74}$& $\mathbf{3.44 \pm 0.78}$& $\mathbf{3.45 \pm 0.95}$& $\underline{\mathbf{3.74 \pm 1.12}}$ \\ \hline
    \end{tabular}
}
    \caption{\textbf{Predicted errors on Non-Gaussian distributions.} Mean Absolute Error (MAE) for Evidental learning with Gaussian outputs, Dropout sampling, Deep Ensembles and Evidental Quantiles. 
    The top scores for each distribution are underlined, and scores within the statistical significance of the evidential quantile models are bolded.
    We use $n_\text{samples}=5$ for the sampling methods.
    The evidential quantiles accurately models the quantiles when the noise is non-Gaussian.}
    \label{tab:non_gaussian_results}
\end{table}

\paragraph{Quantative Benchmarking on UCI datasets}
We benchmark our proposed method on real-world datasets as used by \citet{hernandez2015probabilistic},  \citet{gal2016dropout}, and \citet{lakshminarayanan2017simple}.
We compare the models based on the Tilted loss (TL), the negative log-likelihood (NLL) from the asymmetrical Laplace distributions, and the relative inference speed for the different models compared to our proposed method. 
Table \ref{tab:UCI_benchmark} shows that even though our model is not explicitly trained to optimise these metrics, it remains competitive for both the TL and the NLL compared to the baseline methods.
As mentioned earlier, one of the advantages of our proposed approach is that inference is fast because it only requires one forward pass, as opposed to ensemble and dropout sampling requiring multiple forward passes to sample different weights. Therefore, our approach remains superior with respect to the computational resources needed for inference. Although we do not measure the resources used for training, we note that deep ensembles are $n_\text{samples}$ times more computationally expensive compared to Dropout sampling and Deep evidential quantiles.

\begin{table*}
    \centering
\resizebox{\textwidth}{!}{
    \begin{tabular}{l|rrr|rrr|rrr} \hline
 & \multicolumn{3}{c}{TL} &  \multicolumn{3}{|c|}{NLL} &\multicolumn{3}{c}{Relative inference speed} \\ 

Dataset & \multicolumn{1}{c}{Dropout} &  \multicolumn{1}{c}{Ensemble} &\multicolumn{1}{c|}{Evidential}  & \multicolumn{1}{c}{Dropout} &  \multicolumn{1}{c}{Ensemble} &\multicolumn{1}{c|}{Evidential}  & \multicolumn{1}{c}{Dropout} &  \multicolumn{1}{c}{Ensemble} &\multicolumn{1}{c}{Evidential}  \\ \hline
Boston & $0.58 \pm 0.08$ & $0.53 \pm 0.03$ &  $\underline{\mathbf{0.49 \pm 0.03}}$ &$\mathbf{5.50 \pm 0.23}$ & $\mathbf{5.40 \pm 0.09}$ &  $\underline{\mathbf{5.35 \pm 0.35}}$ &$6.42$ & $4.29$ &  $\underline{\mathbf{1.00}}$  \\
Concrete & $\mathbf{1.10 \pm 0.17}$ & $\underline{\mathbf{1.06 \pm 0.06}}$ &  $\mathbf{1.08 \pm 0.10}$ &$\mathbf{6.78 \pm 0.31}$ & $\underline{\mathbf{6.65 \pm 0.09}}$ &  $\mathbf{6.77 \pm 0.27}$ &$5.85$ & $3.89$ &  $\mathbf{1.00}$  \\
Energy & $0.30 \pm 0.05$ & $0.28 \pm 0.02$ &  $\underline{\mathbf{0.24 \pm 0.02}}$ &$3.91 \pm 0.29$ & $3.75 \pm 0.13$ &  $\underline{\mathbf{3.41 \pm 0.16}}$ &$6.32$ & $4.05$ &  $\underline{\mathbf{1.00}}$  \\
Kin8nm & $0.02 \pm 0.00$ & $0.02 \pm 0.00$ &  $\underline{\mathbf{0.01 \pm 0.00}}$ &$-1.58 \pm 0.08$ & $-1.71 \pm 0.04$ &  $\underline{\mathbf{-1.80 \pm 0.06}}$ &$8.22$ & $4.12$ &  $\underline{\mathbf{1.00}}$  \\
Naval & $\underline{\mathbf{0.00 \pm 0.00}}$ & $\underline{\mathbf{0.00 \pm 0.00}}$ &  $\underline{\mathbf{0.00 \pm 0.00}}$ &$\mathbf{-10.52 \pm 0.41}$ & $\mathbf{-10.58 \pm 0.15}$ &  $\underline{\mathbf{-10.83 \pm 0.36}}$ &$8.63$ & $4.06$ &  $\mathbf{1.00}$  \\
Power-plant & $0.89 \pm 0.02$ & $\mathbf{0.87 \pm 0.01}$ &  $\underline{\mathbf{0.86 \pm 0.01}}$ &$6.44 \pm 0.05$ & $\underline{\mathbf{6.38 \pm 0.02}}$ &  $6.46 \pm 0.04$ &$8.42$ & $4.18$ &  $\underline{\mathbf{1.00}}$  \\
Protein & $\mathbf{0.74 \pm 0.01}$ & $\underline{\mathbf{0.73 \pm 0.00}}$ &  $\mathbf{0.76 \pm 0.04}$ &$5.79 \pm 0.04$ & $\underline{\mathbf{5.75 \pm 0.02}}$ &  $5.89 \pm 0.13$ &$9.41$ & $3.32$ &  $\underline{\mathbf{1.00}}$  \\
Wine & $\underline{\mathbf{0.12 \pm 0.00}}$ & $\underline{\mathbf{0.12 \pm 0.00}}$ &  $\underline{\mathbf{0.12 \pm 0.00}}$ &$\mathbf{2.49 \pm 0.14}$ & $\underline{\mathbf{2.40 \pm 0.03}}$ &  $\mathbf{2.52 \pm 0.17}$ &$6.83$ & $4.20$ &  $\underline{\mathbf{1.00}}$  \\
Yacht & $0.21 \pm 0.04$ & $0.21 \pm 0.01$ &  $\underline{\mathbf{0.17 \pm 0.02}}$ &$2.71 \pm 0.62$ & $2.54 \pm 0.17$ &  $\underline{\mathbf{1.87 \pm 0.41}}$ &$5.85$ & $4.11$ &  $\underline{\mathbf{1.00}}$  \\ \hline
    \end{tabular}
}
    \caption{\textbf{Benchmark regression task}. Tilted loss (TL), negative log-likelihood (NLL), and relative inference speed for Dropout sampling, Deep Ensembles and Evidental Quantiles.
    The top scores for each model and dataset are underlined, while scores (within statistical significance) are bolded.    $n_\text{ensemble}=5$ for sampling baselines. Evidential models outperform baseline models regarding relative inference speed and are competitive with other metrics.}
    \label{tab:UCI_benchmark}
\end{table*}

\paragraph{Disentanglement of uncertainty}
Disentanglement of aleatoric and epistemic uncertainty is critical for applications where it is essential to distinguish between model and observation uncertainty. This is particularly interesting in active learning,  where we wish to avoid acquiring labels for data with high aleatoric uncertainty but are more interested in samples with high epistemic uncertainty \cite{gal2017deep}.
To test the model's ability to disentangle aleatoric and epistemic uncertainty, we construct a dataset of $y=x^3 + \operatorname{Exp}\left(\frac{|x|}{4}\right)$, with $x \in [-4; 4]$. This dataset has high aleatoric uncertainty around the centre ($x=0$). 

\begin{figure}[t]
\centering
\includegraphics[width=0.9\columnwidth]{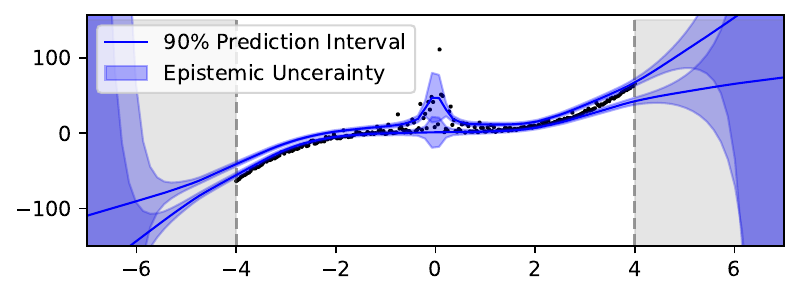}
\includegraphics[width=0.9\columnwidth]{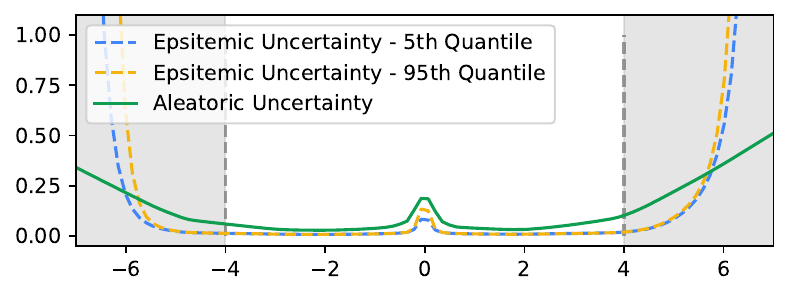}
\caption{\textbf{Disentagled uncertainty} 
Top: fit of the 5th and 95th quantile along with their epistemic uncertainty.
Bottom: Comparison between the Aleatoric and epistemic uncertainties.
Epistemic uncertainty grows more than aleatoric uncertainty on OOD data, reflecting the uncertainty of model parameters outside the training distribution.}
\label{fig:disentagle}
\end{figure}

Figure \ref{fig:disentagle} depicts the epistemic uncertainty for the 5th and 95th quantile and the aleatoric uncertainty for the range $x\in [-7; 7] $. We see that within the data region, the epistemic uncertainty is low, and as we go beyond the support of the data, the epistemic uncertainty increases beyond the aleatoric uncertainty. 

\paragraph{Monocular depth estimation and out-of-distribution detection}
Another key aspect of uncertainty estimation is to detect out-of-distribution (OOD) samples. 
We test the proposed method to detect OOD samples on the complex task of monocular depth estimation.
The problem involves learning a representation of depth directly from an RGB image. This is a challenging learning task, as $y$ is very high-dimensional with predictions at every pixel. 

\begin{figure}[tb]
    \centering
    \includegraphics[width=.9\columnwidth]{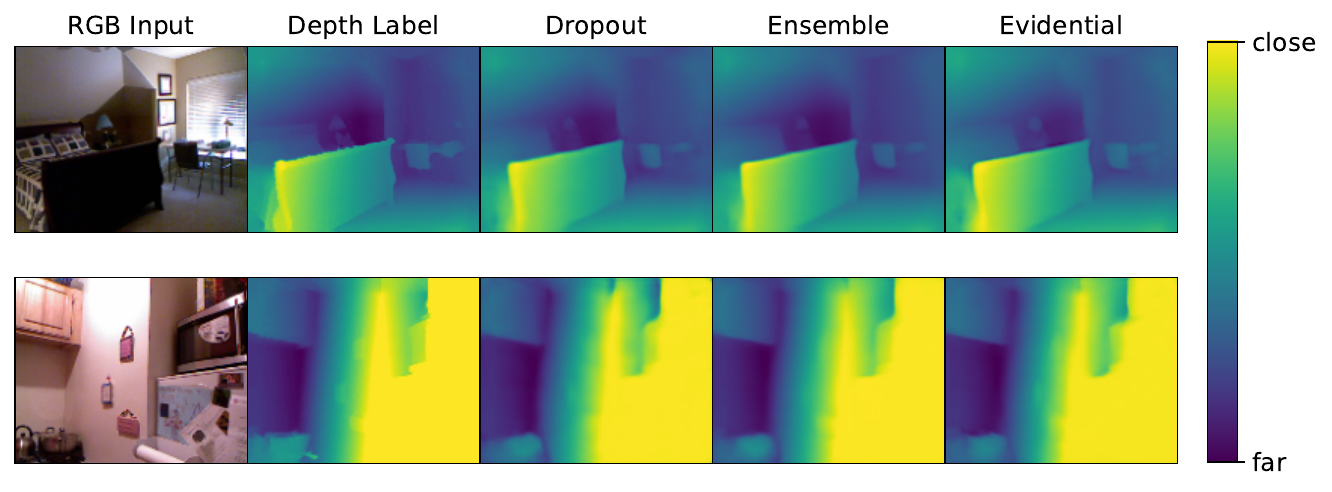}
    \includegraphics[width=.9\columnwidth]{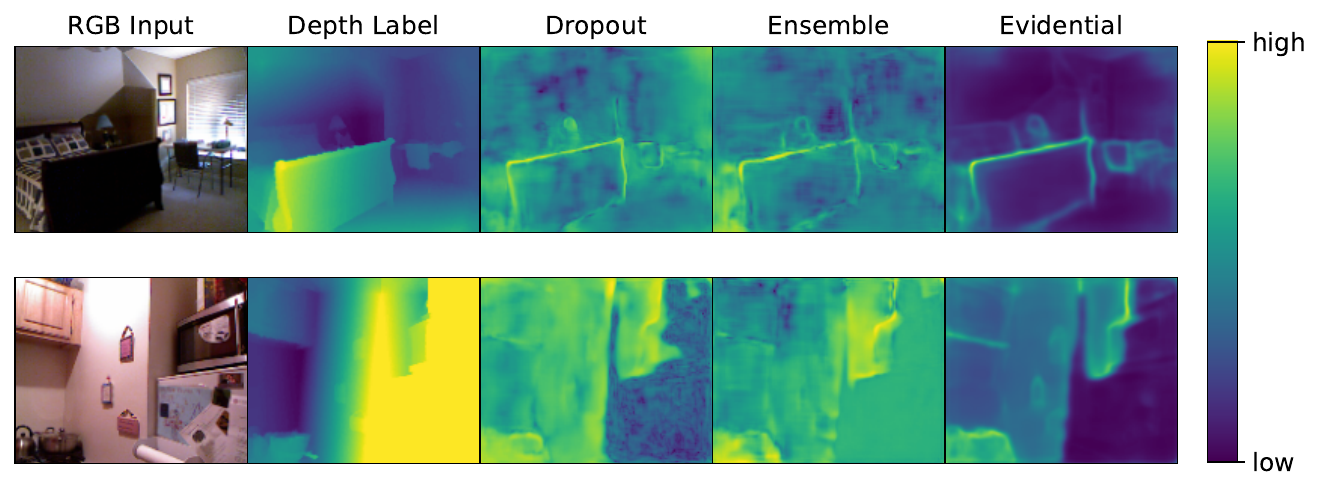}
    \caption{\textbf{Monocular depth estimation } Top shows the predictions of the 95th quantile for two images in the test set. The bottom shows the epistemic uncertainty for the different models for two images in the test set. The evidential epistemic uncertainty is more related to the edges compared to the other methods. }
    \label{fig:mono_detph_in}
\end{figure}

We train our model on the NYU depth v2 Dataset \cite{nyuv2}. The data consist of image pairs of \emph{indoor} scenes (e.g. kitchen, bedrooms, etc.), from RGB to-depth, $H\times W$.
For the case of deep ensembles and dropout sampling, the output is a single $H \times W$ activation map for each quantile.
For the evidential models, the outputs consist of a single $H \times W \times 4$ activation map for each quantile, where the last dimension is ($\gamma, \nu, \alpha, \beta$) for each pixel.
We train a U-Net style NN \cite{ronneberger2015u} with spatial dropout \cite{spatialdropout} to estimate the 5th and 95th quantiles of the depth.
Figure \ref{fig:mono_detph_in} visualises the predicted 95th quantile of the model and the epistemic uncertainty on two images from the test set. 
Ideally, the epistemic uncertainty should capture where the model is making errors, which should be around edges in the case of depth estimation. 
Unlike the Dropout sampling and the deep ensemble, the evidential model captures the depth and provides precise and localised uncertainty predictions.

\begin{figure*}
\centering
\includegraphics[width=.90\textwidth]{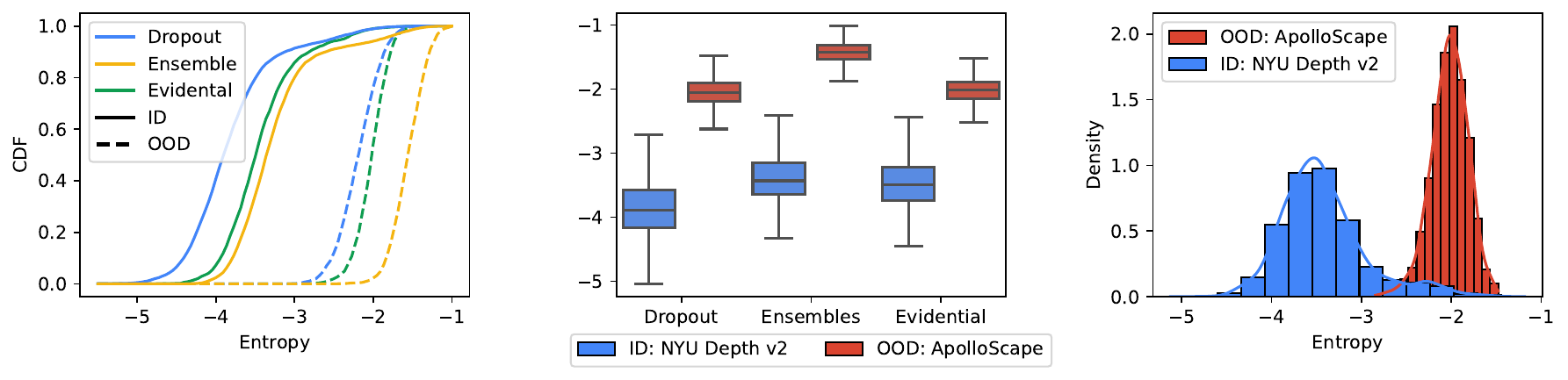} 
\caption{\textbf{Uncertainty on out-of-distribution (OOD) data}. 
The Evidential quantile regression model estimates low uncertainty (entropy) on in-distribution (ID) data and inflates uncertainty for OOD data. Left) Cumulative density function (CDF) of IN and OOD entropy for the 95th quantile for the test sets. Middle: Uncertainty (entropy) comparison across the methods. Right The full histogram of the entropy for the evidential quantile regression.}
\label{fig10:depth_entropy}
\end{figure*}

\begin{figure}
    \centering
    \includegraphics[width=.9\columnwidth]{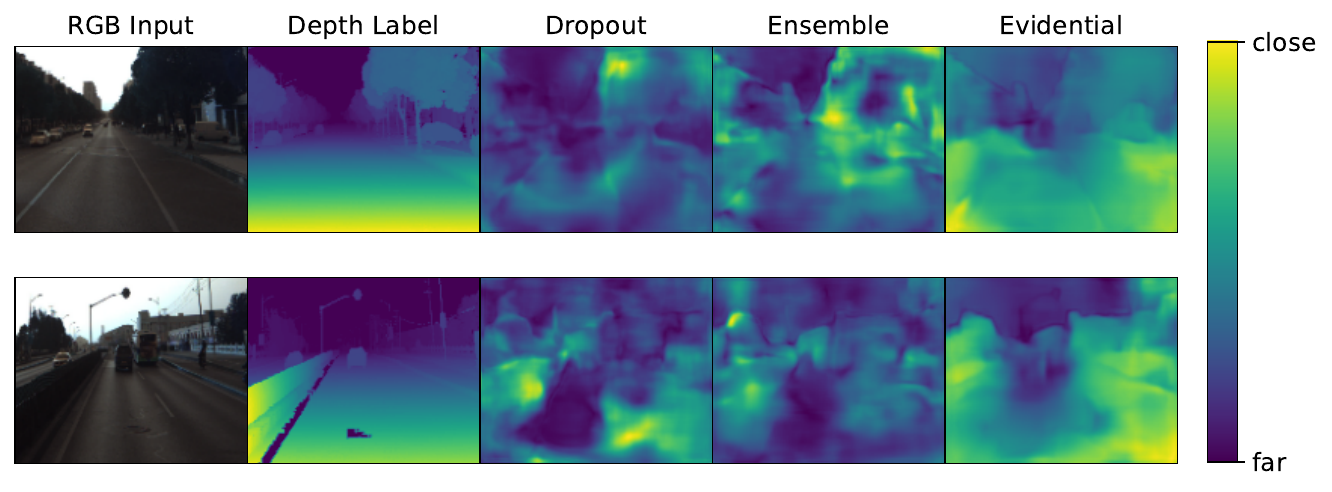}
    \includegraphics[width=.9\columnwidth]{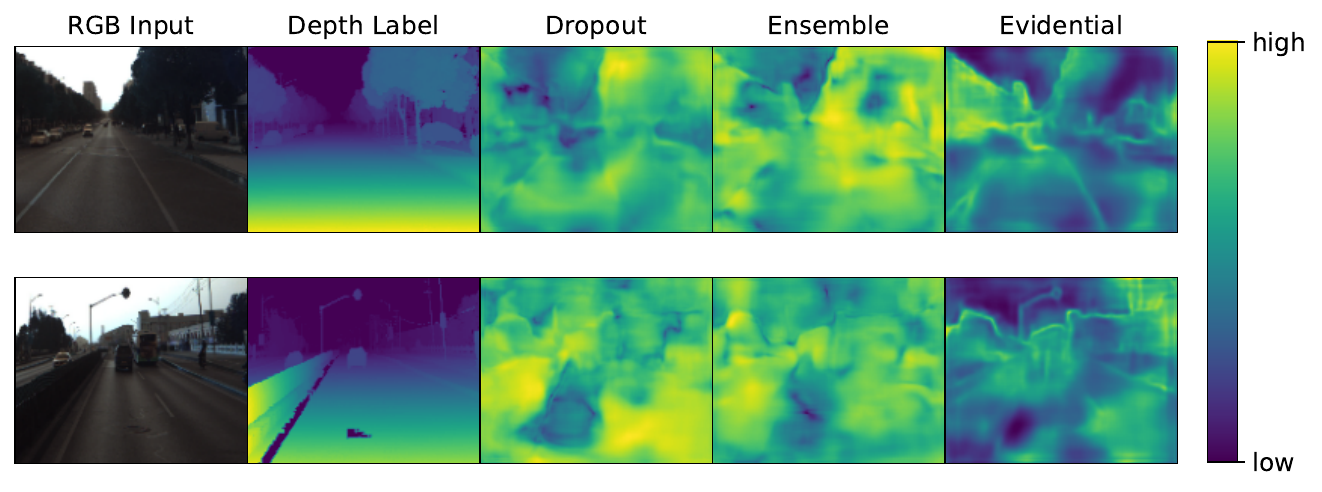}
    \caption{\textbf{Monocular depth estimation for OOD samples}.  
    The top shows the predictions of the 95th quantile for two images in the Apolloscape Dataset. 
    The bottom shows the epistemic uncertainty for the different models.
    The evidential epistemic uncertainty captures the edges of objects.}
    \label{fig:mono_depth_out}
\end{figure}

\paragraph{OOD detection}
To test the ability of the model to capture epistemic uncertainty on OOD, we test our model on images from the ApolloScape dataset \cite{huang2018apolloscape}. Unlike the NYU depth dataset, the ApolloScape dataset contains scenes from outside driving.
Figure \ref{fig:mono_depth_out} visualises the predicted 95th quantile of the model and the epistemic uncertainty for the two images in the ApolloScape dataset.
Again, we notice that the evidential model captures the depth of the image while still providing clear epistemic uncertainty of the depth estimations.

To detect OOD samples, we predict the epistemic uncertainty on the test set on both IN distribution samples from the NYU depth dataset and OOD samples from the ApolloScape dataset and record the average entropy for each image in the datasets.
Figure \ref{fig10:depth_entropy} left shows each method's cumulative density function (CDF) on the test set. Clearly, all methods show a distinct shift in the entropy of the CDFs between the in-distribution (ID) and the out-of-distribution (OOD). Our evidential model performs equally well as the two baselines.
Figure \ref{fig10:depth_entropy} middle shows the boxplot of the entropy distributions again between the IN and OOD samples. It separates the two distributions with a few overlapping samples. 
On the right in Figure \ref{fig10:depth_entropy}, we show the complete entropy distribution for the evidential model.
These results show that the proposed evidential model captures the increased uncertainty in OOD data, and the uncertainty estimates are comparable to other methods of estimating epistemic uncertainty.

\section{Related works}
\paragraph{Bayesian Quantile Regression}
Quantile regression models were introduced as an alternative to the traditional regression to the mean \cite{koenerk1978regression}.
Quantile regression widely applies to many regression-based tasks where the aleatoric uncertainty is non-Gaussian. For example, in traffic forecasting  \cite{rodrigues2020beyond, Nguyen_Quanz_2021}, retail demand forecasting \cite{sun2023neural}, censored regression \cite{huttel2022Modeling},and distributional reinforcement learning \cite{Dabney_Rowland_Bellemare_Munos_2018}.
Traditionally the Bayesian quantile regression models assume the Asymmetrical Laplace distribution of the quantiles \cite{yu20201bayesian}, and it can be extended to the mixture representation in Equation \ref{eq:scalar_mixture_model} \cite{kotz2001laplace}.
For sampling-based approaches such as Gibbs sampling, \cite{kozumi2011gibbs}, starts with the mixture representation with the Normal Inverse Gamma prior over the mean and noise parameters \cite{feng2015bayesian, chu2023bayesian}, as it allows for approximate bayesian inference of the quantile regression problem. 
Bayesian quantile regression has also been extended to Gaussian processes with expectation-maximisation \cite{Boukouvalas2012Gaussian} variational inference \cite{Abeywardana_Ramos_2015}, or expectile optimisation \cite{Torossian2020BayesianQA}.
Our work addresses some limitations of fitting Bayesian quantile regression on large datasets by not using sampling-based techniques to model the epistemic uncertainty.

\paragraph{Bayesian Deep Learning and Evidential Learning}
Bayesian Deep learning aims to infer the posterior distribution over the network weights during training and use the posteriors for uncertainty estimation. 
In general, exact Bayesian inference of the weights is intractable due to the size of the parameter space, and the functional form of the models does not lend itself to exact integration \cite{kingma2015variational, blundell2015weight}. 
The posteriors can then be approximated with variational inference \cite{blundell2015weight}, dropout sampling \cite{gal2016dropout, gal2017concrete} and Deep ensembles \cite{lakshminarayanan2017simple, pearce2018high, pearce2020uncertainty}.
However, these approaches are limited by the requirements of multiple forward passes to compute the weight posteriors, making them computationally expensive to run in practical applications. 
As an alternative to inferring the posteriors over the weights in BNNs, 
one can view deep learning as an evidence-acquisition process \cite{sensoy2018evidental}.  
In the evidential framework, each observation is a sample that, during training, supports a higher-order evidential distribution that we are trying to approximate. 
The evidential distributions describe the epistemic and aleatoric uncertainty of the model \cite{pandey2023learn}.
Several works use this approach, such as \emph{prior network}, which places Dirichlet priors over the discrete classifications tasks \cite{sensoy2018evidental}.
However, they struggle with generalisation (particularly without using out-of-distribution training data) \cite{malinin2018predictive, hafner2020noise} and modelling aleatoric uncertainty accurately \cite{gurevich2020gradient}.
These issues have been largely addressed for regression tasks by \citet{amini2020deep}. 
However, one of the limitations of prior work on deep evidential regression is that the aleatoric uncertainty is assumed to be Gaussian. Our work addresses this limitation by estimating the quantiles of the target distribution as opposed to just the mean and variance.

\paragraph{Limitations}
While our method does provide some advantages over current techniques, one of the main limitations to our work and evidential regression is tuning the regularisation strength of the evidence.
We believe it should be optimised for each use case and that it is one of the biggest drawbacks of deep evidential learning.
In addition, the proposed method does not necessarily outperform the compared baseline but can achieve similar performance with less computational resources for inference.

\section{Conclusion}
This work proposes a novel and scalable deep Bayesian quantile regression model by placing evidential priors over the likelihood function.
The model has the flexibility to model aleatoric uncertainty for non-Gaussian distribution.
We have demonstrated that the method can disentangle aleatoric and epistemic uncertainty, detect out-of-distribution samples and scale to large computer vision tasks.






\newpage

\bibliography{aaai24}

\newpage
\appendix
\section{Deriviations}
\subsection{Derivitation of the evidential distribution}
\label{sec:deriviation}
Here derive the t-distribution from the marginal likelihood under the NIG prior. 

\begin{equation}
    p\left(y \mid \boldsymbol{m}\right)=\frac{p\left(y \mid \boldsymbol{\theta}, \boldsymbol{m}\right) p(\boldsymbol{\theta} \mid \boldsymbol{m})}{p\left(\boldsymbol{\theta} \mid y, \boldsymbol{m}\right)}\, ,
\end{equation} 
is equal to the maximum likelihood of the t-distribution. We omit indications of observation $i$ and the dependency of $q$ for readability.

\begin{equation}
        p\left(y \mid \boldsymbol{m}\right)= \int_{\sigma_i=0}^{\infty} \int_{\mu=-\infty}^{\infty} p\left(y \mid \boldsymbol{\theta}_i\right) p\left(\boldsymbol{\theta} \mid \boldsymbol{m}\right) \mathrm{d} \mu \mathrm{d} \sigma
\end{equation}
\begin{equation}
        p\left(y \mid \boldsymbol{m}\right)= \int_{\sigma=0}^{\infty} \int_{\mu=-\infty}^{\infty} p\left(y \mid \mu, \sigma\right) p\left(\mu, \sigma \mid \boldsymbol{m}\right) \mathrm{d} \mu \mathrm{d} \sigma 
\end{equation}

$p(y \mid \mu, \sigma)$ is the likelihood for a Gaussian distribution with mean $\mu+\tau z$ and variance $\omega \sigma z$, and $p(\mu, \sigma \mid \boldsymbol{m}$ is the Normal Inverse Gamma prior.

\begin{equation}    
\begin{split}
    p\left(y \mid \boldsymbol{m}\right)= &\int_{\sigma, \mu}\sqrt{\frac{1}{2\pi \omega \sigma z}} \exp \left(-\frac{y-\mu-\tau z}{2\omega\sigma z}\right) \frac{\beta^\alpha \sqrt{v}}{\Gamma(\alpha) \sqrt{2 \pi \sigma}}\\
    &\quad \left(\frac{1}{\sigma}\right)^{\alpha+1} \exp \left\{-\frac{2 \beta+v(\gamma-\mu)^2}{2 \sigma}\right\}  d\mu d\sigma
\end{split}
\end{equation}

\begin{equation}
        p\left(y \mid \boldsymbol{m}\right)= \int_{\sigma=0}^\infty
    \frac{\sqrt{\nu } \beta ^{\alpha } \sigma ^{-\alpha -2} \sqrt{\frac{1}{\omega z}} e^{-\frac{2 \beta +\frac{\nu  (\gamma +\tau z-y)^2}{\omega \nu  z+1}}{2 \sigma }}}{\sqrt{2 \pi } \Gamma (\alpha ) \sqrt{\frac{\tau \nu  z+1}{\omega \sigma  z}}}
    d\sigma 
\end{equation}
\begin{equation}
\label{eq:final_proof}
\small
       p\left(y \mid \boldsymbol{m}\right) = \frac{\sqrt{\nu } \beta ^{\alpha } \Gamma \left(\alpha +\frac{1}{2}\right) \sqrt{\frac{1}{\omega z}} \left(\frac{2 \omega \nu  z+2}{2 \beta  (\omega \nu  z+1)+\nu  (\gamma +\tau z-y)^2}\right)^{\alpha +\frac{1}{2}}}{\sqrt{2 \pi } \Gamma (\alpha ) \sqrt{\frac{1}{\omega z}+\nu }}
\end{equation}
After marginalizing, we show that this is a t-distribution
We match up the likelihood function for the t-distribution with Equation \ref{eq:final_proof}.
The t-distribution is defined as, 
\begin{equation}
\small
    f(y | \alpha_{\text{st}}, \mu_{\text{st}}, \sigma_{\text{st}}) = \frac{\Gamma(\frac{\alpha_{\text{st}}+1}{2})}{\Gamma(\frac{\alpha_{\text{st}}}{2})}\frac{1}{\sqrt{\pi \alpha_{\text{st}} \sigma_{\text{st}}^2}}\left(1+\frac{(y-\mu_{\text{st}})^2}{\alpha_{\text{st}}\sigma_{\text{st}}^2}\right)^{-\frac{\alpha_{\text{st}}+1}{2}}\, .
\end{equation}

Firstly we rewrite Equation \ref{eq:final_proof} such that, 
\begin{equation}
\small
 p(y\mid \boldsymbol{m}) = \frac{\Gamma(\frac{a+1}{2})}{\Gamma(\frac{a}{2})}\frac{1}{\pi a}\sqrt{\frac{a \nu}{2\omega z}} \beta^{a/2}\frac{\left[ \frac{2(1+\omega z \nu)}{(y-\gamma-\tau z)^2\nu+2\beta(1+\omega z\nu)}\right]^{\frac{a+1}{2}}}{\sqrt{\frac{1}{\omega z +\nu}}}
\end{equation}
From here we see that $\mu_{\text{st}} = \gamma + \tau z$ and $\alpha_{\text{st}}=2a$.
We now match $\frac{1}{\sqrt{\sigma_{\text{st}}^2}}$, to the rest of the likelihood for the t-distribution.

\begin{equation}
   \frac{1}{\sqrt{\sigma_{\text{st}}^2}} = \sqrt{\frac{\alpha \nu}{2\omega z}}\frac{\beta^{(\alpha+1)/2}}{\sqrt{\beta}}\frac{\left[\frac{2(1+\omega z\nu)}{(\gamma-\mu_{\text{st}})^2\nu+2\beta(1+\omega z\nu)} \right]^{\frac{\alpha+1}{2}}}{\sqrt{\frac{1}{\omega z}\sqrt{1+\omega z\nu}}}
\end{equation}

\begin{equation}
    = \sqrt{\alpha\nu}\frac{\left[\frac{2\beta(1+\omega z\nu)}{(y-\mu_{\text{st}})^2\nu+2\beta(1+\omega z\nu)} \right]^{\frac{\alpha+1}{2}}}{\sqrt{2\beta(1+\omega z\nu)}}
\end{equation}

\begin{equation}
    = \frac{\sqrt{\alpha\nu}}{\sqrt{2\beta(1+\omega z\nu)}}\left[ 1+\frac{\nu(y-\mu_{\text{st}})^2}{2\beta(1+\omega z\nu)}\right]^{\frac{a+1}{2}}
\end{equation}

\begin{equation}
    = \frac{1}{\sqrt{\sigma_{\text{st}}^2}}\left(1+\frac{(y-\mu_{\text{st}})^2}{\alpha \sigma_{\text{st}}^2} \right)^{-\frac{\alpha+1}{2}}
\end{equation}

\begin{equation}
    \frac{1}{\sqrt{\sigma_{\text{st}}^2}} = \sqrt{\frac{\alpha\nu}{(2\beta(1+\omega z\nu)}}
\end{equation}

\begin{equation}
    \frac{1}{\alpha\sigma_{\text{st}}^2} = \frac{\nu}{(2\beta(1+\omega z\nu)}
\end{equation}

\begin{equation}
    \frac{1}{\sigma_{\text{st}}^2} = \frac{\alpha \nu}{(2\beta(1+\omega z\nu)}
\end{equation}

\begin{equation}
    \sigma_{\text{st}}^2=\frac{2\beta(1+\omega z\nu)}{\alpha \nu}
\end{equation}

In conclusion 
\begin{equation}
\begin{split}
    p\left(y\mid \boldsymbol{m}\right)
    &= \int_{\sigma=0}^{\infty} \int_{\mu=-\infty}^{\infty} p\left(y \mid \mu, \sigma\right) p\left(\mu, \sigma \mid \boldsymbol{m}\right) \mathrm{d} \mu \mathrm{d} \sigma  \\
    &= \text{St}(y;\gamma+ \tau z, \frac{2\beta(1+\omega\nu z)}{\nu \alpha}, 2\alpha) \, .
\end{split}
\end{equation}

\section{Experimental Setups and model architectures}
\label{app:experiments}
All the code and data can be found at \textit{github.com/fbohu/evidential-quantile-regression}
All results with dense models are the average across 20 runs, and all scores include the average plus minus two times the standard deviation.
All the data pre-processing steps follow the same procedure outlined by \citet{amini2020deep}
All dense models are run on a single-core Intel Xeon Processor 2650v4.
For the monocular depth estimation, we run the experiment 1 time, 
All convolutional models are run on Tesla NVIDIA GV100 with 32gb ram.
The final version will include a citation to the specific cluster used.

\subsection{Qualitative Results}
For the qualitative results presented in Figure \ref{fig:quali_eli}, we use a densely connected neural network with 128 hidden units for three layers for all methods. We add 0.1 dropout \cite{srivastava2014dropout} for each model and use the Leaky ReLU activation function.
We use the Adam optimiser with a learning rate of $3\cdot10^{-3}$ with a batch size of 128 \cite{kingma2014adam}.
We train the models on 1000 samples from the exponential distribution $ y = x^3 + \varepsilon, \quad \varepsilon \sim \operatorname{Exp}\left(\frac{1}{4|x|+0.2} \right)$ for 1000 epochs with early stopping for 250 epochs.
For the evidential regularisation, we use $\lambda = 0.3$

\subsection{Ablation study of the evidence regularisation strength}
\begin{figure}
    \centering
    \includegraphics[width=0.9\columnwidth]{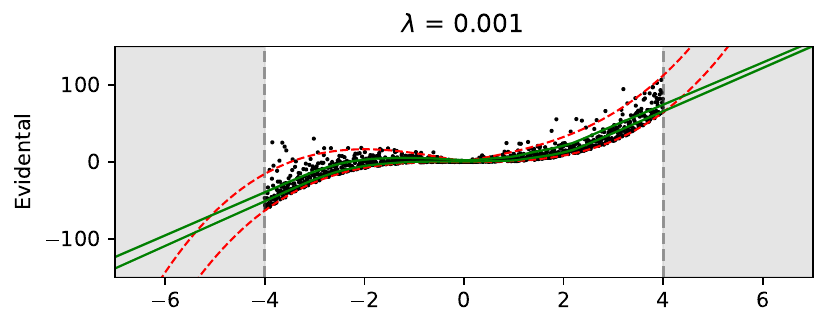}
    \includegraphics[width=0.9\columnwidth]{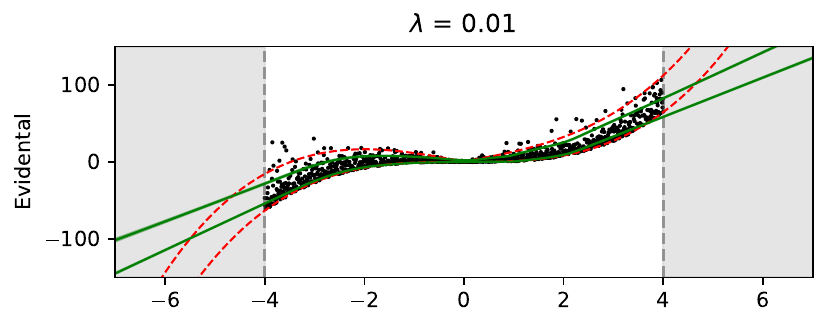}
    \includegraphics[width=0.9\columnwidth]{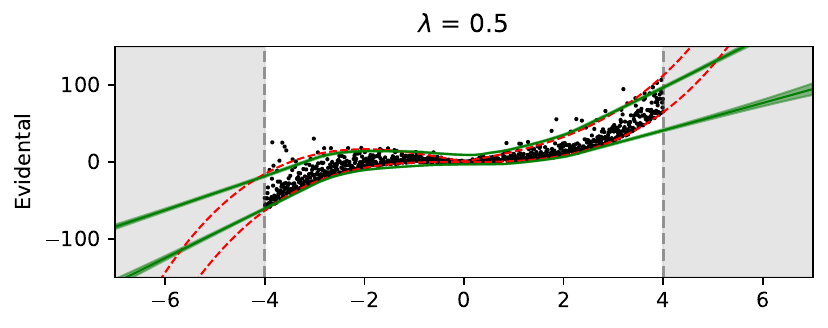}
    \includegraphics[width=0.9\columnwidth]{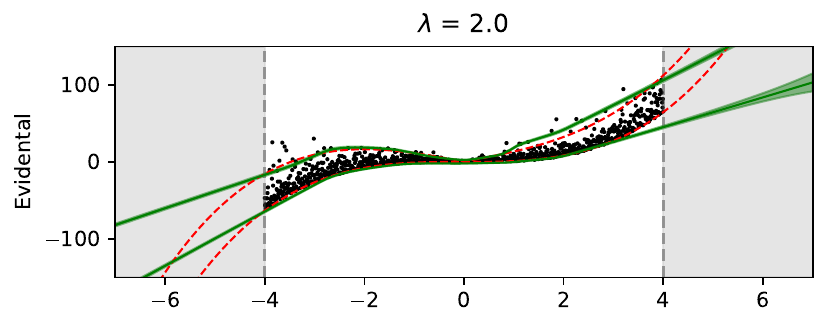}
    \includegraphics[width=0.9\columnwidth]{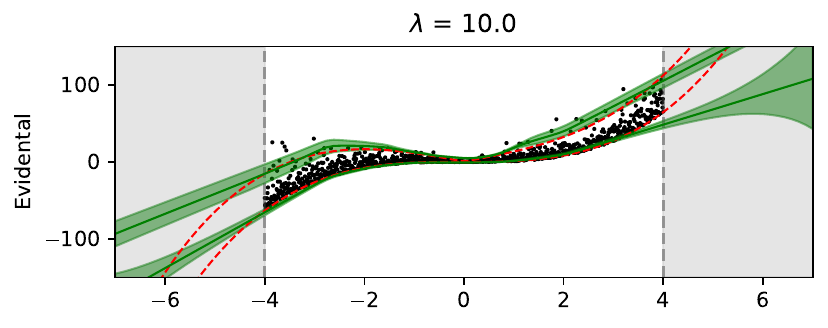}
    \caption{\textbf{Aleatoric Uncertainty for different regularization strength}}
    \label{fig:abb_ale}
\end{figure}

\begin{figure}
    \centering
    \includegraphics[width=0.9\columnwidth]{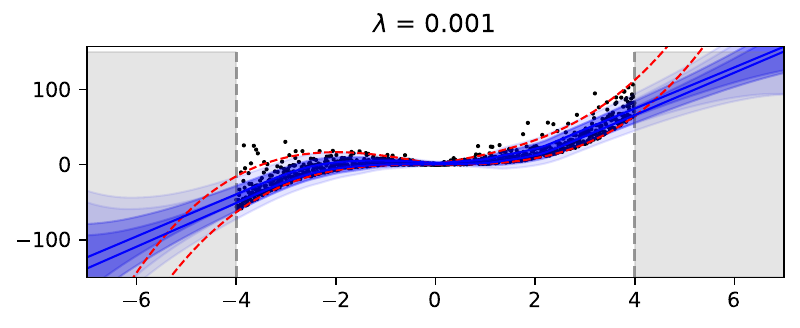}
    \includegraphics[width=0.9\columnwidth]{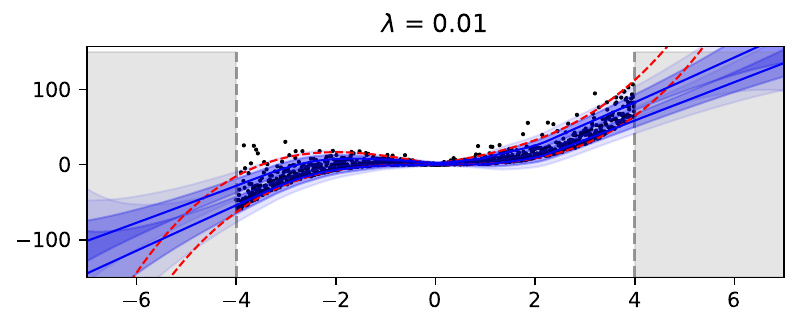}
    \includegraphics[width=0.9\columnwidth]{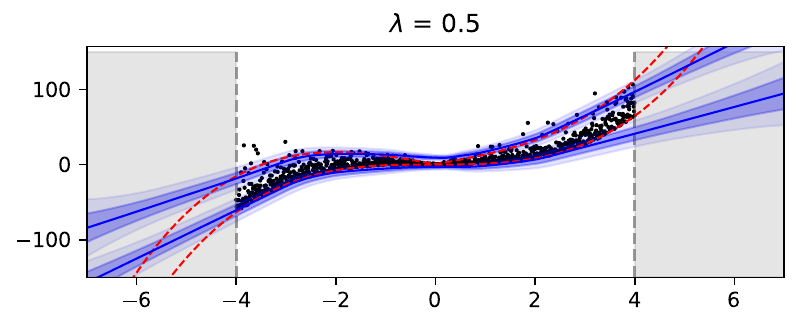}
    \includegraphics[width=0.9\columnwidth]{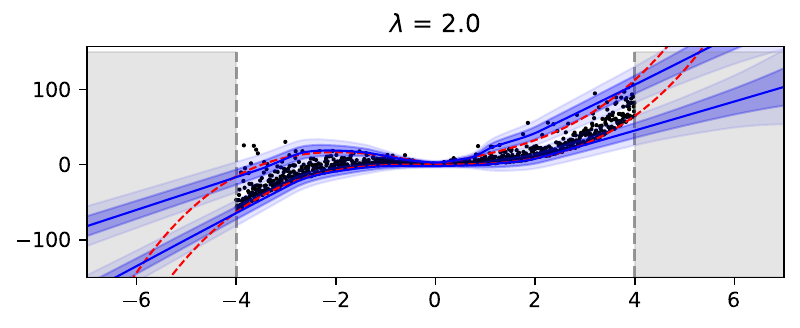}
    \includegraphics[width=0.9\columnwidth]{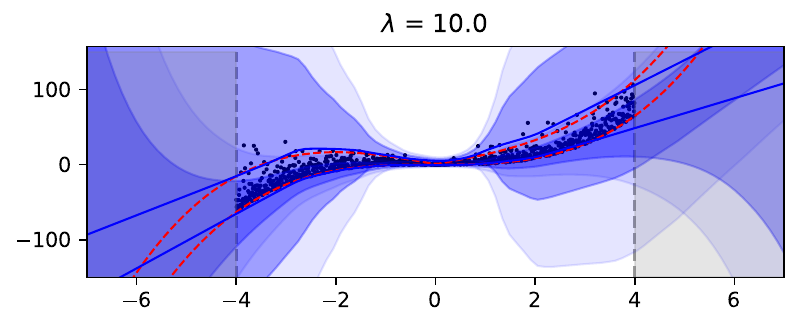}
    \caption{\textbf{Epistemic Uncertainty for different regularization strength}}
    \label{fig:abb_evi}
\end{figure}

The following experiment demonstrates the importance of the regularisation strength on the evidence in Equation \ref{eq:total_loss}.
Figure \ref{fig:abb_ale} and Figure \ref{fig:abb_evi} provide the qualitative results on the epistemic and aleatoric uncertainty, with different regularisation coefficients, $\lambda$.
We show that the estimated quantiles tend towards the mean instead of the desired quantiles as we decrease the regulariser.
As we increase the strength, the epistemic and aleatoric uncertainty is inflated.
We also see that as we increase the regularization strengths, the better we approximate the quantiles at the expense of the uncertainty calibration. 

\subsection{Beyond the Gaussian Assumption}
To test the models on the non-Gaussian distribution presented in Table \ref{tab:non_gaussian_results}, we train a densely connected neural network with 256 hidden units and three layers for all methods.
We keep the activation function and the dropout rates the same as above. 
Again, we use the Adam optimizer with a learning rate of $5 \cdot10{-3}$. 
We train the model on 5000 samples from the noise distributions and use early stopping over 50 epochs. We keep the batch size 32.
We find that $\lambda=0.5$ works quite well across the UCI datasets experiments, but naturally, this parameter should be optimised for each experiment.

\subsection{Quantitative Results on UCI datasets}
For the benchmarking results on the UCI datasets in Table \ref{tab:UCI_benchmark}, we train a densely connected neural network with 128 hidden units and two layers.
We keep the activation function and the optimizer the same.
We use Bayesian optimization on a validation set to identify the optimal parameters for each method for each method's dropout, learning rate, and batch size. Therefore, they are not constant across the methods. 
We use the expected improvement to acquire new configurations of the hyperparameters.
The search areas are

\begin{table}[h]
    \centering
    \begin{tabular}{l |r| r}
        Hyperparameter & Lower & Upper  \\ \hline
        Dropout & 0.1 & 0.5 \\
        Learning rate & $1 \cdot10{-5}$ & $5\cdot10{-3}$ \\
        Batch size & 16 & 128 \\ \hline
    \end{tabular}
    \caption{Search space for the hyperparameters of the UCI benchmark dataset.}
    \label{tab:my_label}
\end{table}

\subsection{Disentanglement of uncertainty}
To test the model to disentangle the aleatoric and epistemic uncertainty, as shown in Figure \ref{fig:disentagle}, we train a densely connected neural network with 128 hidden units and three layers. 
We keep the activation function and dropout rates as above.
We add $\ell_2$ regularization to the model with $\lambda_{\ell_2}= 10^{-4}$. 
We make the $\lambda$ for the evidential regulariser $\lambda=0.3$.
We train the model on 5000 samples from the distribution with a batch size 32. 
We apply early stopping over 50 epochs.

\subsection{Monocular Depth Estimation.}
For the Monocular Depth estimation experiments, we use the same setup as in \cite{amini2020deep} with the NYU-Depth-v2 Dataset  \cite{nyuv2}.
We use the same train and test split as \cite{amini2020deep}, where the depth labels are the proportional inverse of the actual depth. 
This is common for depth estimation as it ensures that far-away objects result in numerically stable models. 
We use the same splits for testing and testing as in \cite{amini2020deep}

The model architectures have a U-net \cite{ronneberger2015u} architecture similar to the one used in \cite{amini2020deep}. 
The network consists of five convolutional and pooling blocks down and back up. 
The input data has the shape (160, 128) with 3 input channels for the RGB, while the target only has a single depth feature map. 
All models are trained with spatial dropout (p=0.1) over the convolutional blocks \cite{spatialdropout}.
Our evidential models have four output feature maps, one for each parameter $\gamma$, $\nu$, $\alpha$, and $\beta$, as described in Section 4.

All models were trained with the following hyperparameters: batch size 32, Adam optimiser with a learning rate $5 \cdot 10^{-5}$, for 500 epochs. 
All models use spatial dropout with 0.1.
Evidential model had $\lambda=0.5$. 

\subsection{Quantitative results on Monocular Depth Estimation}
Table \ref{tab:mono_results_simple}, summarises the qualitative results for the monocular depth estimation in terms of the Tilted loss (TL) and Negative log likelihood (NLL) of the asymmetrical Laplace distribution
\begin{table}[h]
    \centering
    \begin{tabular}{l|rrr} \hline
        Model & TL & NLL & Parameters \\ \hline
        Dropout  & 0.0099 &  2.33 & 7.846.756\\
        Ensemble & 0.0088 & -4.70 & 392.33.780\\
        Evidential & 0.0098 & 0.56& 7.846.960 \\ \hline
    \end{tabular}
    \caption{\textbf{Depth Estimation performance metrics}, The tilted loss the negative log-likelihood and the number of parameters used in each model.}
    \label{tab:mono_results_simple}
\end{table}
We did not tune any hyperparameters for this experiment and use standard parameters as \cite{amini2020deep}.

\end{document}